\documentclass[10pt,twocolumn,letterpaper]{article}

\usepackage{iccv}
\usepackage{times}
\usepackage{epsfig}
\usepackage{graphicx}
\usepackage{amsmath}
\usepackage{amssymb}
\usepackage{authblk}

\usepackage{subcaption}     
\usepackage{multirow}
\usepackage{booktabs}       
\usepackage{algorithm2e}
\RestyleAlgo{ruled}
\SetKwComment{Comment}{/* }{ */}

\usepackage[breaklinks=true,bookmarks=false]{hyperref}

\iccvfinalcopy 

\def\iccvPaperID{****} 

\ificcvfinal\fi

\begin{document}

\title{Leveraging Classic Deconvolution and Feature Extraction in Zero-Shot Image Restoration}


\author[1,4]{Tomáš Chobola\thanks{tomas.chobola@helmholtz-muenchen.de}}
\author[2]{Gesine Müller}
\author[3]{Veit Dausmann}
\author[3]{Anton Theileis}
\author[3]{Jan Taucher}
\author[2]{Jan Huisken}
\author[4]{Tingying Peng\thanks{tingying.peng@helmholtz-muenchen.de}}
\affil[1]{Technical University of Munich, Munich, Germany}
\affil[2]{Georg-August-University Göttingen, Göttingen, Germany}
\affil[3]{GEOMAR Helmholtz Centre for Ocean Research Kiel, Kiel, Germany}
\affil[4]{Helmholtz AI, Helmholtz Munich - German Research Center for Environmental Health, Neuherberg, Germany}

\maketitle
\ificcvfinal\thispagestyle{empty}\fi

\begin{abstract}
    Non-blind deconvolution aims to restore a sharp image from its blurred counterpart given an obtained kernel. Existing deep neural architectures are often built based on large datasets of sharp ground truth images and trained with supervision. Sharp, high quality ground truth images, however, are not always available, especially for biomedical applications. This severely hampers the applicability of current approaches in practice. In this paper, we propose a novel non-blind deconvolution method that leverages the power of deep learning and classic iterative deconvolution algorithms. Our approach combines a pre-trained network to extract deep features from the input image with iterative Richardson-Lucy deconvolution steps. Subsequently, a zero-shot optimisation process is employed to integrate the deconvolved features, resulting in a high-quality reconstructed image. By performing the preliminary reconstruction with the classic iterative deconvolution method, we can effectively utilise a smaller network to produce the final image, thus accelerating the reconstruction whilst reducing the demand for valuable computational resources. Our method demonstrates significant improvements in various real-world applications non-blind deconvolution tasks.
\end{abstract}

\section{Introduction}

Image deconvolution is a classic problem in computer vision and imaging sciences. It aims to recover a sharp image $x$ out of a blurred and noisy representation
\begin{equation}\label{eq:degrade}
    y = x*k+b,
\end{equation}
where $k$ is the blur kernel, $b$ is additive noise and $*$ denotes the convolution operator. Traditionally, methods that solve this problem are split into two steps: first, the blur kernel $k$ is obtained either through estimation or imaging system calibration. Then, image restoration is performed (i.e., non-blind deconvolution). Given the reduction or removal of high frequencies caused by convolving an image with a blur kernel and unknown noise caused by measurement errors, image deconvolution is essentially a challenging ill-posed inverse problem.

Recently, multiple studies of deep neural networks designed for sharp image recovery have been conducted \cite{dong2020deep,8237753,zhang2017learning2,jin2017noise,eboli2020end,ren2018deep,dong2021learning}. While those methods are able to achieve impressive performance on computer vision benchmarks \cite{levin, lai_dataset, 6528301, nah2017deep}, the pre-trained models are not directly applicable to scientific imaging given the arising domain gap, and therefore it decreases their generalisation performance. Additionally, re-training or fine-tuning is often not possible given the sparsity of ground truth sharp images in scientific imaging fields such as medical or biological microscopy \cite{sparse_data, zollner2020image, Jadhav2020}. This problem cannot be alleviated by using images from other scientific fields due to the image diversity, which results in large domain shifts and gaps between different types of microscopy images. 

To address the limitations arising from insufficient training data, various deep self-supervised techniques have been developed for deblurring and denoising tasks, as evidenced in previous works \cite{ulyanov2018deep, gandelsman2019double, selfdeblur, chen2022self}. However, these methods primarily cater to general image restoration tasks and do not account for the distinctive features and complexities inherent in scientific images. Scientific images often exhibit unique noise patterns, intricate structures, and specialized imaging modalities, necessitating tailored restoration approaches. Therefore, the development of specialized deep self-supervised methods becomes imperative to effectively tackle restoration challenges specific to scientific images. Moreover, it is important to emphasize that existing methods often neglect the consideration of computational costs and model size. In the context of scientific image restoration, these aspects hold crucial importance due to limited computational resources and the need for efficiently processing large quantities of data. Therefore, the development of deep self-supervised methods that also take into account computational efficiency and model compactness is highly beneficial in real-life scientific applications.

Given the reasons stated above, in many scientific fields deconvolution is still handled by classic algorithms \cite{Richardson:72, landweber, wiener1949extrapolation} that are fast, but known to be prone to degradation and noise amplification. To improve the image quality, total variation (TV) has become the standard for noise reduction and regularisation in image reconstruction, but in recent years its limitations such as loss in textures, changes in smooth intensity or patchy artefacts have been reported \cite{cai2012image, guo2014new, tomographic2016}. 

The core contributions of this paper can be summarised as follows: 
\begin{enumerate}
    \item We propose CiDeR\footnote{\href{https://github.com/ctom2/cider}{https://github.com/ctom2/cider}} (\textbf{C}lass\textbf{i}c \textbf{De}convolution and Feature Extraction for Zero-Shot Image \textbf{R}estoration), a self-supervised approach for non-blind deconvolution embedding the Richardson-Lucy algorithm, a classic method for image deconvolution, into a deep learning model.
    \item We integrate the continuity prior as a form of Hessian regularisation into the loss function, reducing the noise in the resulting sharp image while preserving the underlying image structures.
    \item We apply our proposed CiDeR to real microscopy images, addressing a variety of image restoration tasks across different microscopy modalities, achieving enhanced visual quality and improved image details.
\end{enumerate}

\begin{figure*}[h!]
    \centering
    \includegraphics[width=.8\linewidth]{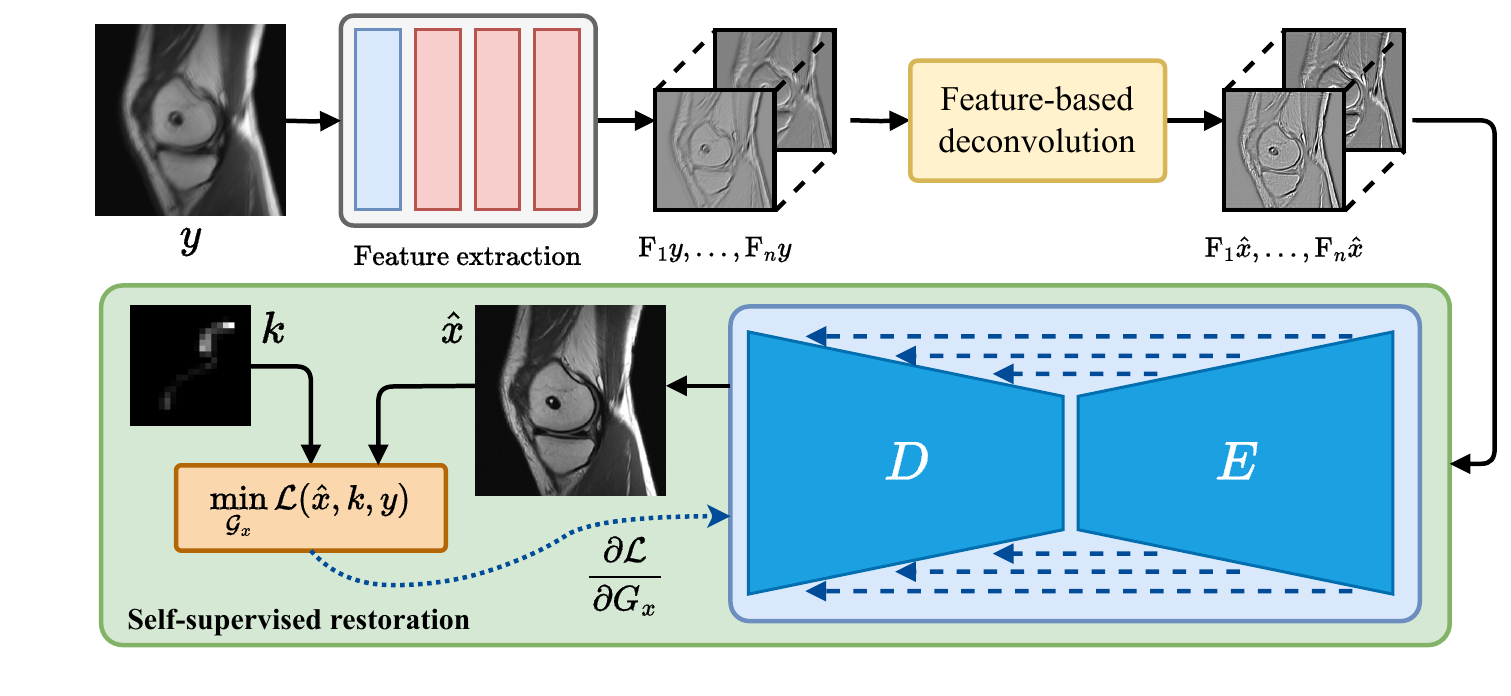}
    \caption{The architecture of CiDeR, a novel non-blind deconvolution model that integrates the Richardson-Lucy algorithm into a zero-shot optimisation framework.}
    \label{fig:architecture}
\end{figure*}

\section{Related Work}

In this section, we discuss various classic and deep image deconvolution methods and their shortcomings. We also mention methods used in specific scientific domains where image enhancement plays a key role.

\subsection{Classic Deconvolution Algorithms} 

One of the earliest methods for non-blind deconvolution are the Wiener filter \cite{wiener1949extrapolation} and Richardson-Lucy \cite{Richardson:72} that impose assumptions that image noise follows Gaussian and Poisson distributions, respectively. Besides, Landweber iteration \cite{landweber} is another popular algorithm for sharp image recovery, which is a special case of gradient descent. While those methods are fast and simple in their implementations, they are prone to degradation and noise amplification. To improve the reconstructed image quality, multiple optimisation methods employing various priors, such as Laplacian, hyper-Laplacian \cite{levin2007image, krishnan2009fast} or total variation regularisation \cite{rudin1992nonlinear, wang2008new}, have been developed. Yuan \textit{et al.} \cite{yuan2008progressive} proposed an inter-scale and intra-scale non-blind deconvolution method to obtain fine details while suppressing image artefacts. Manually designed priors usually rely on statistical features of the natural images, which is not suitable for all image types and can lead to inference problems.

\subsection{Deep Deconvolution Models}

Recently, deep models have been used for image deconvolution problems \cite{zhang2017learning, zhang2017learning2, schuler2013machine, xu2014deep} to learn the features needed for image restoration instead of engineering them. Gong \textit{et al.} \cite{gong2020learning} incorporated deep neural networks into a fully parameterised gradient descent scheme to learn an implicit image prior. The combination of the Wiener filter and CNNs into a deep supervised method was proposed by Dong \textit{et al.} \cite{dong2020deep}, where the deconvolution is performed on learned image features and not in the image space to improve the level of detail and noise suppression. Jointly learning spatially-variant data and regularisation terms within the MAP framework, as proposed by Dong \textit{et al.} \cite{dong2021learning}, better captures the properties of clear images. While those methods achieve impressive performance on benchmark computer vision datasets, they require large quantities of ground truth data to train that are not accessible in biological or medical imaging sciences. Additionally, using models trained on images from a domain that differs from the target domain introduces a domain shift, which could significantly reduce the performance in the target domain and therefore the overall model usability. 

Moreover, there has been a growing interest in the development of deep restoration methods that leverage classic deconvolution algorithms for volumetric data reconstruction, aiming to enhance processing efficiency \cite{li2022incorporating, chobola2023lucyd}. These approaches combine the principles of classical deconvolution with advancements in deep learning and optimisation techniques to efficiently restore volumetric datasets.

The concept of deep image prior (DIP) \cite{ulyanov2018deep,gandelsman2019double} has been successfully applied to denoising, super-resolution and deconvolution \cite{selfdeblur}. Chen \textit{et al.} \cite{chen2022self} applied an ensemble of DIP models for image deblurring. These generative models alleviate the need for training datasets as they are optimised only given an input image and a kernel, hence being more flexible and adaptive. While these methods perform well in general image restoration, they often overlook the distinctive features and attributes of scientific images, demanding customised solutions to effectively tackle their specific challenges. It is worth noting that these methods do not take into account the importance of limited computational resources. Therefore, developing tailored solutions for scientific image restoration remains essential.

\subsection{Domain Specific Deconvolution} 

Increasing the spatial resolution through deconvolution is a widely studied problem in the microscopy imaging field. Zhao \textit{et al.} \cite{zhao2022sparse} proposed a processing pipeline for fluorescence microscopy consisting of background removal, upsampling reconstruction based on sparsity and continuity priors and iterative deconvolution with Richardson-Lucy \cite{Richardson:72}. Image deconvolution algorithms in domains such as optical, electron and X-ray microscopy \cite{WATANABE2002191,Sibarita2005,1628876, Ehn2016,zhao2022sparse} rely on a specific imaging setup or image features characteristic for the target domain. This prevents the use of such algorithms in different fields and makes their window of applicability extremely narrow. 

\section{Proposed Method}

We propose a combination of a classic non-blind iterative deconvolution algorithm and deep learning. The objective of the method is to restore a representation of the sharp image based purely on the blurry input image and the appropriate kernel in a \textit{self-supervised} manner. By embedding the Richardson-Lucy algorithm \cite{Richardson:72} into the method we omit the need for using an extensively large neural network to produce the sharp image.

\subsection{Richardson-Lucy Algorithm}

The Richardson-Lucy algorithm is a restoration technique widely used in various fields, including astronomy \cite{white1994image}, microscopy \cite{zhao2022sparse}, and medical imaging \cite{dell2010modified}. The primary objective of the method is to enhance the quality of images degraded by blur and noise during the acquisition process (following Equation \ref{eq:degrade}). The algorithm works iteratively and at each iteration it estimates an intermediate restored image $x^{(i)}$, where $i$ represents the iteration number. The update rule is as follows,
\begin{equation}\label{eq:rl}
    x^{(i)}=x^{(i-1)}\cdot\left(\frac{y}{x^{(i-1)}*k}*k^\top\right).
\end{equation}
The restoration begins with an initial estimate of the original image $x^{(0)}$ which can be set to the observed image $y$ or any other reasonable initialisation. The algorithm iterates until a convergence criterion is met or a predefined number of iterations is reached. One of the significant advantages of the Richardson-Lucy algorithm is its ability to handle deconvolution even in cases where the noise characteristics are unknown or complex. 

\subsection{Feature Deconvolution With Richardson-Lucy}

Similarly to \cite{dong2020deep}, we address the deconvolution task by utilising deep features instead of the conventional image space representation. The standard image space deconvolution often proves to be insufficient in effectively removing artifacts and restoring fine details in degraded images \cite{dong2018learning, shan2008high}. To overcome these limitations, we exploit deep features, which capture high-level abstract information learned from convolutional neural networks, to guide the restoration process more effectively. 

Incorporating a set of linear filters $\{f_j\}_{j=1}^n$ generated through deep neural networks, that are designed to extract feature information from the degraded input, we can establish the relationship between $y$, $x$, and $k$ within the feature space as follows,
\begin{equation}
    \mathbf{F}_j\mathbf{y}=\mathbf{KF}_j\mathbf{x} + \mathbf{F}_j\mathbf{b},
\end{equation}
where $\mathbf{F}_j$, $\mathbf{K}$, $\mathbf{y}$, $\mathbf{x}$, and $\mathbf{b}$ denote the matrix/vector forms of $f_j$, $k$, $y$, $x$, and $b$. Then, the Richardson-Lucy algorithm can be directly applied to the latent features $\{\mathbf{F}_j\mathbf{y}\}$ as follows,
\begin{equation}
    \mathbf{F}_j\mathbf{x}^{(i)}=\mathbf{F}_j\mathbf{x}^{(i-1)}\left(\mathbf{K}^\top\frac{\mathbf{F}_j\mathbf{y}}{\mathbf{KF}_j\mathbf{x}^{(i-1)}}\right).
\end{equation}
Upon completing the iterations, we acquire the features of the desired latent clear image denoted as $\{{\mathbf{F}_j\hat{\mathbf{x}}}\}$.

\subsection{Self-Supervised Image Synthesis}

In order to produce high-quality images, we adopt the DIP assumptions \cite{ulyanov2018deep, gandelsman2019double, selfdeblur} and utilise a generative network $G_x$. This network takes the features of the latent clear image as input and it has the structure of an asymmetric Autoencoder \cite{ronneberger2015u} with skip-connections \cite{ulyanov2018deep}. Unlike traditional DIP approaches, where the generative network must possess significant modeling capacity to generate images with rich textures and salient structures, we can reduce the overall model size. This is because the deconvolution objective has already been performed in the latent space, and thus, the purpose of the network shifts to image synthesis rather than performing the deconvolution process itself.

To impose regularisation on the generator, our primary approach involves utilising the structural similarity index (SSIM) \cite{wang2004image}. In conjunction with the SSIM, we incorporate a regularisation term $\mathcal{R}(\cdot)$ aimed at capturing important image priors as follows,
\begin{equation}
    \mathcal{L}(x,k,y)=\alpha\cdot\mathcal{L}_\text{SSIM}(x*k,y)
    +\lambda\cdot\mathcal{R}(x).
\end{equation}
This combination facilitates an effective means of enforcing structural coherence and enhancing the overall quality of the generated outputs. The integration of SSIM and the image prior regularisation empowers our generator to produce more visually appealing and contextually meaningful results, ensuring the preservation of essential characteristics present in the original data. Specifically, we adopt the Hessian prior, which operates effectively in the image space,
\begin{equation}
    \mathcal{R}(z)=\lVert{z_{xx}}\rVert_1 + \lVert{z_{yy}}\rVert_1 + 2\lVert{z_{xy}}\rVert_1,
\end{equation}
where $z_i$ denotes the second-order partial derivatives of $z$ along the $x$ and $y$ axes. The purpose of the regularisation is to reduce the noise artifacts arising from the deconvolution steps and to preserve the underlying image structure. Figure \ref{fig:architecture} illustrates the full model architecture used to synthesise the restored image.

\subsection{Enhancing Microscopy Image Restoration}
\label{subsection:background_sparsity}

In the context of real microscopy image restoration, we initiate the process with a pre-processing step involving background removal. This step helps in enhancing the clarity of the images and preparing them for further restoration. Additionally, to optimise the synthesis process, we modify the loss function by introducing sparsity as an additional prior, alongside the Hessian continuity prior. The inclusion of the sparsity prior is well-founded, as it has demonstrated significant improvements in the reconstruction of super-resolution microscopy images \cite{nature_sparse}. By leveraging both the Hessian and sparsity priors, our approach aims to achieve superior results by enhancing the visual quality of the microscopy images, and providing more accurate representations of the underlying structures.

\subsubsection{Background Removal} 

To address background interference in microscopy images, such as light diffraction or scattering effects, we utilise the modified iterative wavelet transform method \cite{wavelet}, as described by Zhao \textit{et al.} \cite{nature_sparse}. This approach is applied to improve the reliability and visual quality of the subsequent image restoration processes. 

We use the residual image arising from setting the values over the mean value of the original image $y$ to zero to estimate the background as follows: \textit{(i)} The background is iteratively estimated from the lowest frequency wavelet bands related to the input image using 2D Daubechies-6 wavelet filters to decompose the signal up to the 7$^{\text{th}}$ level. \textit{(ii)} An inverse wavelet transform on the lowest band of the frequency information to the spatial domain is performed to prevent the unintended removal of important information. The result is then combined with $\sqrt{y}/2$ into a single image whose pixels consist of the minimum of those two. \textit{(iii)} The output of the previous step is then used as the input in the next iteration. Following \cite{nature_sparse}, we set the number of iterations to 3 to estimate the background. The final estimated background is then subtracted from the original input image $y$.

\subsubsection{Sparsity Prior}

To cater specifically to the unique characteristics of microscopy data, we adapt our approach by augmenting the loss function $\mathcal{L}$ with sparsity regularisation. The sparsity-aware loss function ensures that the reconstruction process not only preserves continuity but also encourages the final result to align with the sparse characteristics typically exhibited in microscopy data. It is important to note that this sparsity regularisation is deliberately included for microscopy image restoration, and may not be utilised in general image reconstruction scenarios. This approach allows us to achieve superior synthesis results by yielding more accurate representations of the underlying content.

To quantify sparsity, we utilise the $\ell_1$ norm on the synthesized image, which leads to the following adapted loss function,
\begin{align}
    \mathcal{L}(x,k,y)&=\alpha\cdot\mathcal{L}_\text{SSIM}(x*k,y)\\
    &+\lambda\cdot\mathcal{R}(x)+\beta\cdot\lVert{x\rVert}_1.
\end{align}

\begin{algorithm}
    \caption{CiDeR image synthesis}\label{alg:two}
    \textbf{Input:} $\mathbf{y},\mathbf{k}$\\
    \textbf{Output:} $\hat{\mathbf{x}}$\\
    $\mathbf{F}_{1:n}\mathbf{y}\gets\text{feature\_extractor}(\mathbf{y})$\\
    $\mathbf{F}_{1:n}\hat{\mathbf{x}}\gets\text{Richardson-Lucy}(\mathbf{F}_{1:n}\mathbf{y})$\\
    \For{$i\gets 1$ \KwTo $T$}{
        $\mathbf{x}\gets G_x^{t-1}(\mathbf{F}_{1:n}\hat{\mathbf{x}})$\\
        Compute the gradient w.r.t. $G_x$\\
        Update $G_x^t$ using the NAdam algorithm \cite{dozat2016incorporating}\\
    }
    $\hat{\mathbf{x}}=G_x^T(\mathbf{F}_{1:n}\hat{\mathbf{x}})$
\end{algorithm}

\begin{figure*}[h!]
    \captionsetup[subfigure]{font=scriptsize,labelfont=scriptsize,{justification=centering}}
    \centering
    \begin{subfigure}[t]{0.8\textwidth}
        \centering
        \includegraphics[width=\linewidth]{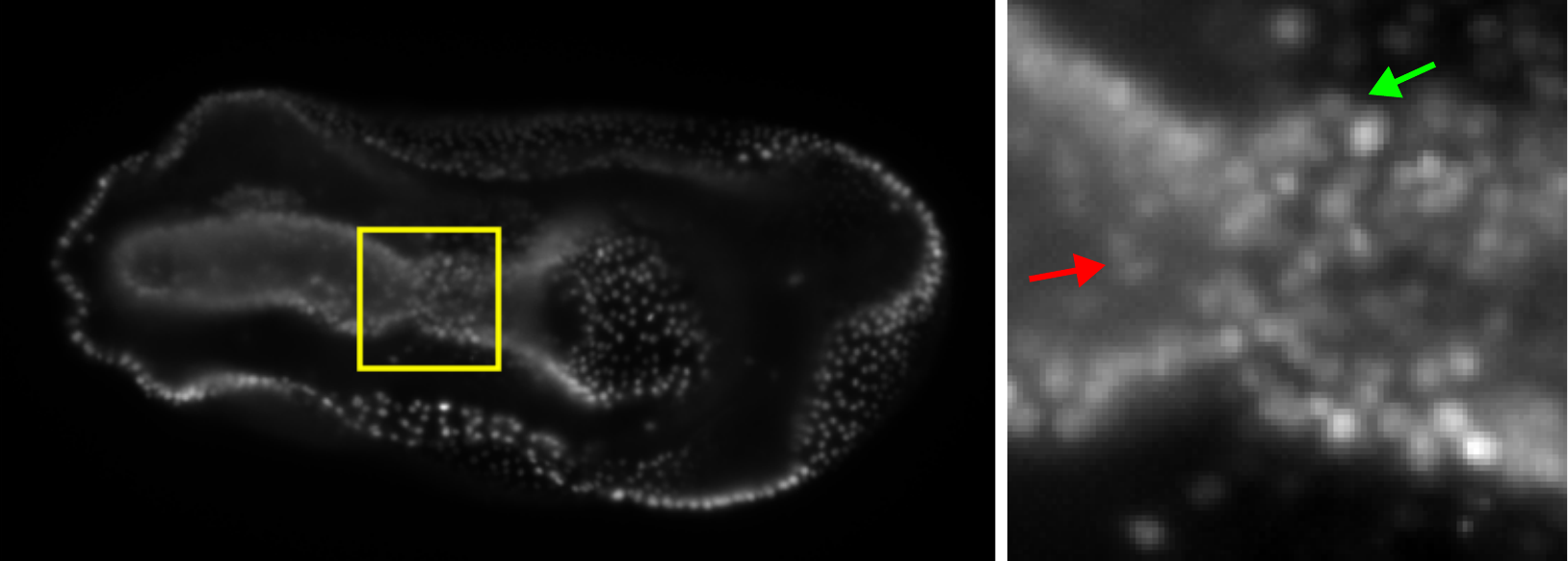}
        \caption{}
    \end{subfigure}
    \\
    \begin{subfigure}[t]{0.8\textwidth}
        \centering
        \includegraphics[width=\linewidth]{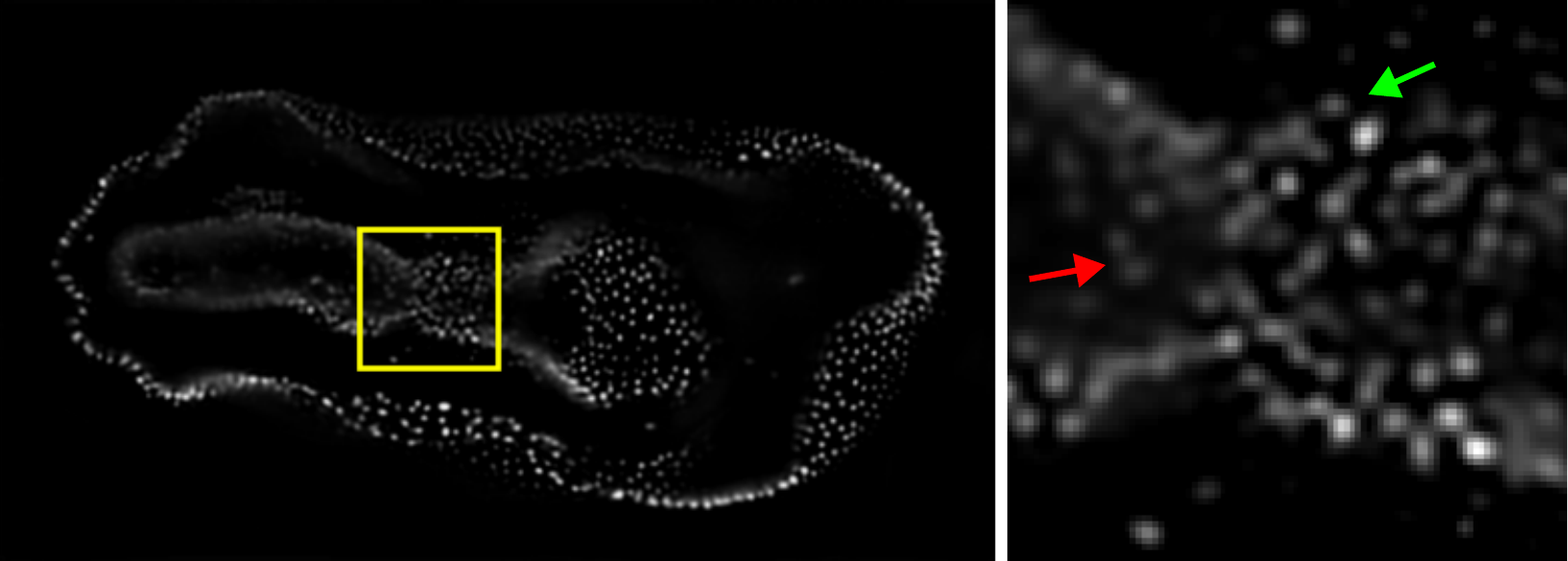}
        \caption{}
    \end{subfigure}
    \caption{A comparison between the original raw light-sheet microscopy image of a starfish (a) and the deconvolved output obtained using CiDeR (b) reveals significant improvements. The deconvolved image effectively reduces haze and light diffraction artifacts within the sample, resulting in clearer and more accurate structures that were previously distorted due to image degradation.}
    \label{fig:starfish}
\end{figure*}

\section{Experimental Results}

We implemented CiDeR using PyTorch \cite{paszke2017automatic}. For extracting features from the degraded images, we employed a pre-trained model from \cite{dong2020deep}, that was fine-tuned on images from the Berkeley segmentation \cite{martin2001database} and Waterloo Exploration \cite{ma2016waterloo} datasets. This feature extractor consists of one convolutional layer and three residual blocks \cite{he2016deep}, enabling the extraction of $n=16$ features.

The generator $G_x$ for the synthesis of the final image is optimised over $T=3000$ iterations using the NAdam algorithm \cite{dozat2016incorporating}. The initial learning rate is set as $0.01$ and is decayed by multiplying $0.5$ when reaching $2,000$, $2,300$ and $2,700$ iterations.

\subsection{Results with Real Degradation}

\begin{figure}[h!]
    \captionsetup[subfigure]{font=scriptsize,labelfont=scriptsize,{justification=centering}}
    \centering
    \begin{subfigure}[t]{0.15\textwidth}
        \centering
        \includegraphics[width=\linewidth]{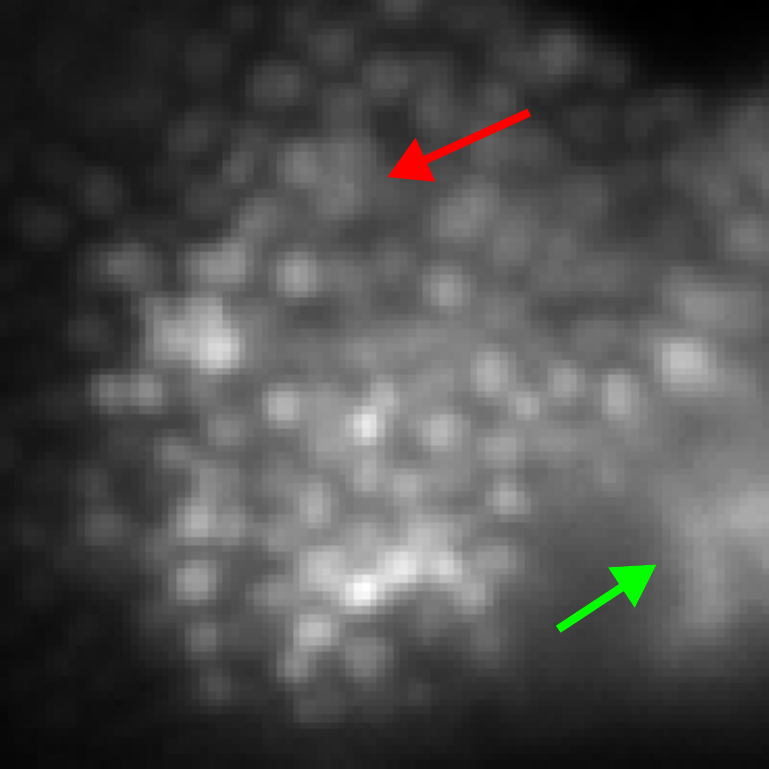}
        \caption{}
    \end{subfigure}
    \begin{subfigure}[t]{0.15\textwidth}
        \centering
        \includegraphics[width=\linewidth]{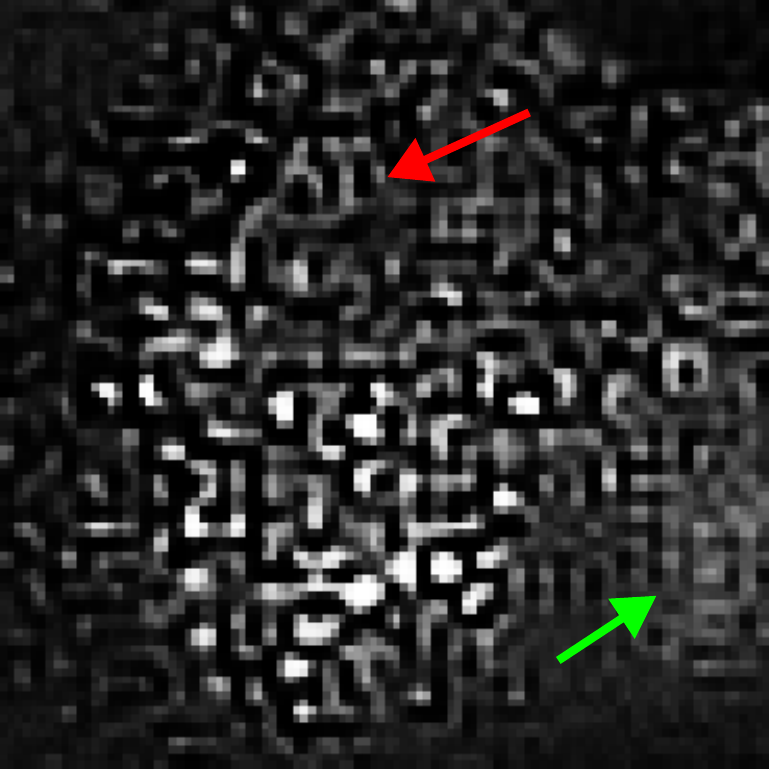}
        \caption{}
    \end{subfigure}
    \begin{subfigure}[t]{0.15\textwidth}
        \centering
        \includegraphics[width=\linewidth]{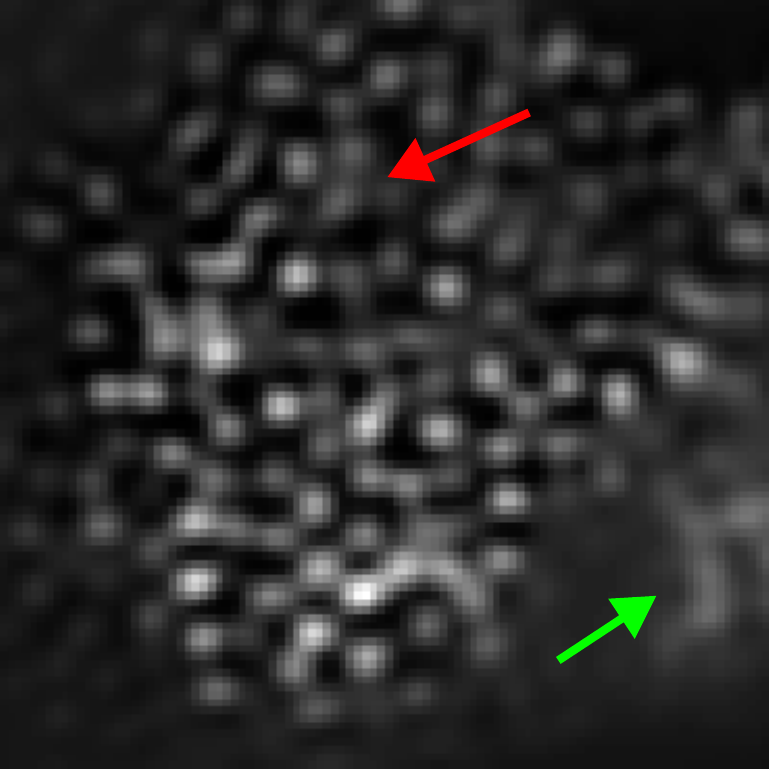}
        \caption{}
    \end{subfigure}
    \caption{Visual comparison involving three images: (a) the raw image of a starfish embryo captured by a light-sheet microscope, (b) the deconvolution of a background-subtracted image using SelfDeblur \cite{selfdeblur}, and (c) the image restoration result obtained using CiDeR.}
    \label{fig:starfish2}
\end{figure}

We assess the performance of CiDeR on two real microscopy modalities: light-sheet and underwater microscopy with simulated microscopy kernels. In Figure \ref{fig:starfish}, we present the primary comparison between the raw light-sheet image and the restoration result obtained using CiDeR. The restoration with CiDeR enhances the separation between cells, elevates the contrast, and improves the overall image quality. Additionally, Figure \ref{fig:starfish2} provides a comparison between CiDeR and SelfDeblur \cite{selfdeblur} on an additional light-sheet image. Both methods employ a self-supervised approach for image restoration, but SelfDeblur is susceptible to overfitting, leading to the introduction of artifacts and image distortions that degrade the final result. In contrast, the incorporation of more fitting image priors in the optimisation objective of CiDeR prevents such degradation, yielding a more faithful and visually pleasing restoration outcome.

\begin{figure}[h!]
    \captionsetup[subfigure]{font=scriptsize,labelfont=scriptsize,{justification=centering}}
    \centering
    \begin{subfigure}[t]{0.45\textwidth}
        \centering
        \includegraphics[width=\linewidth]{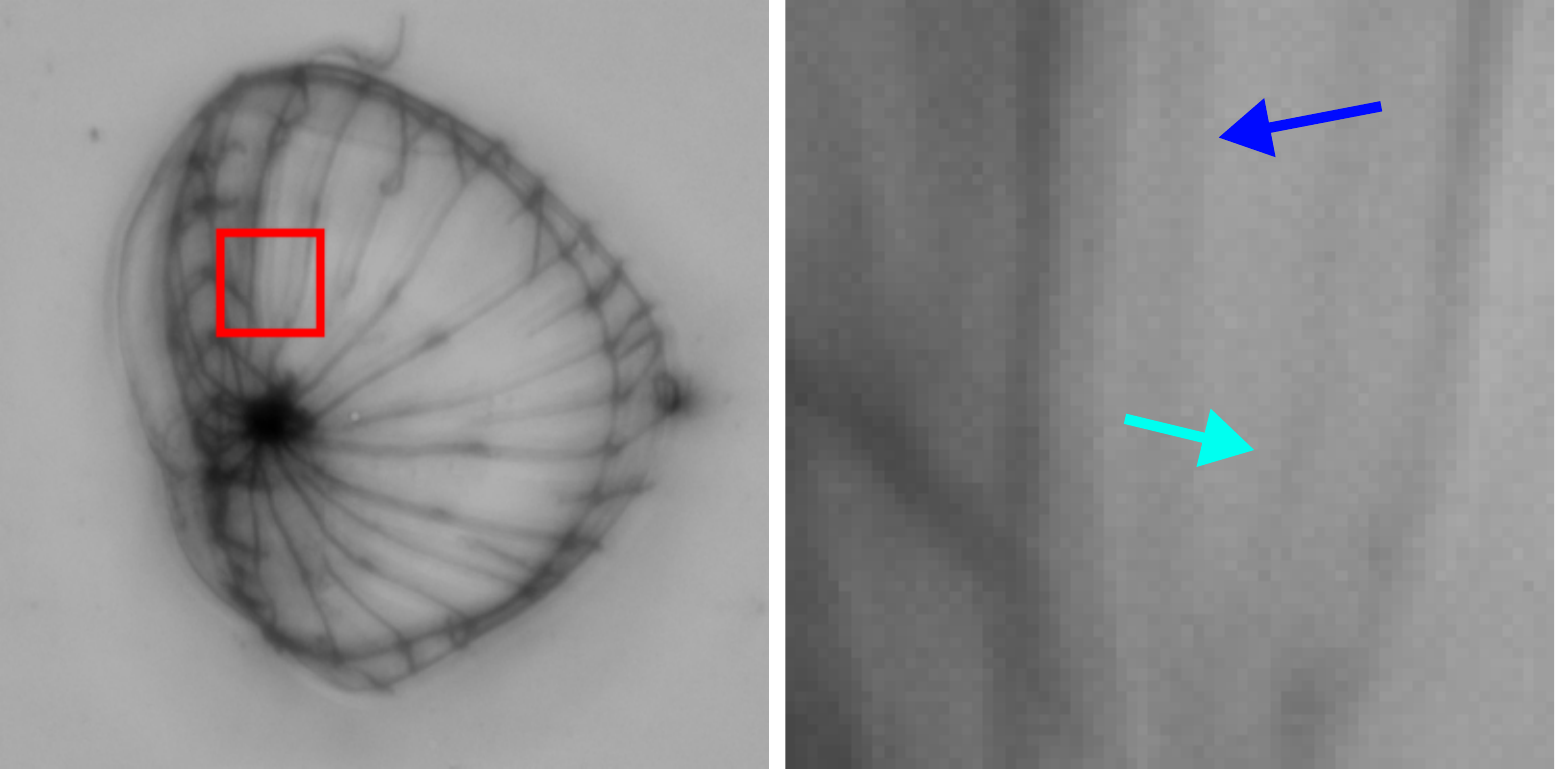}
        \caption{}
    \end{subfigure}
    \\
    \begin{subfigure}[t]{0.45\textwidth}
        \centering
        \includegraphics[width=\linewidth]{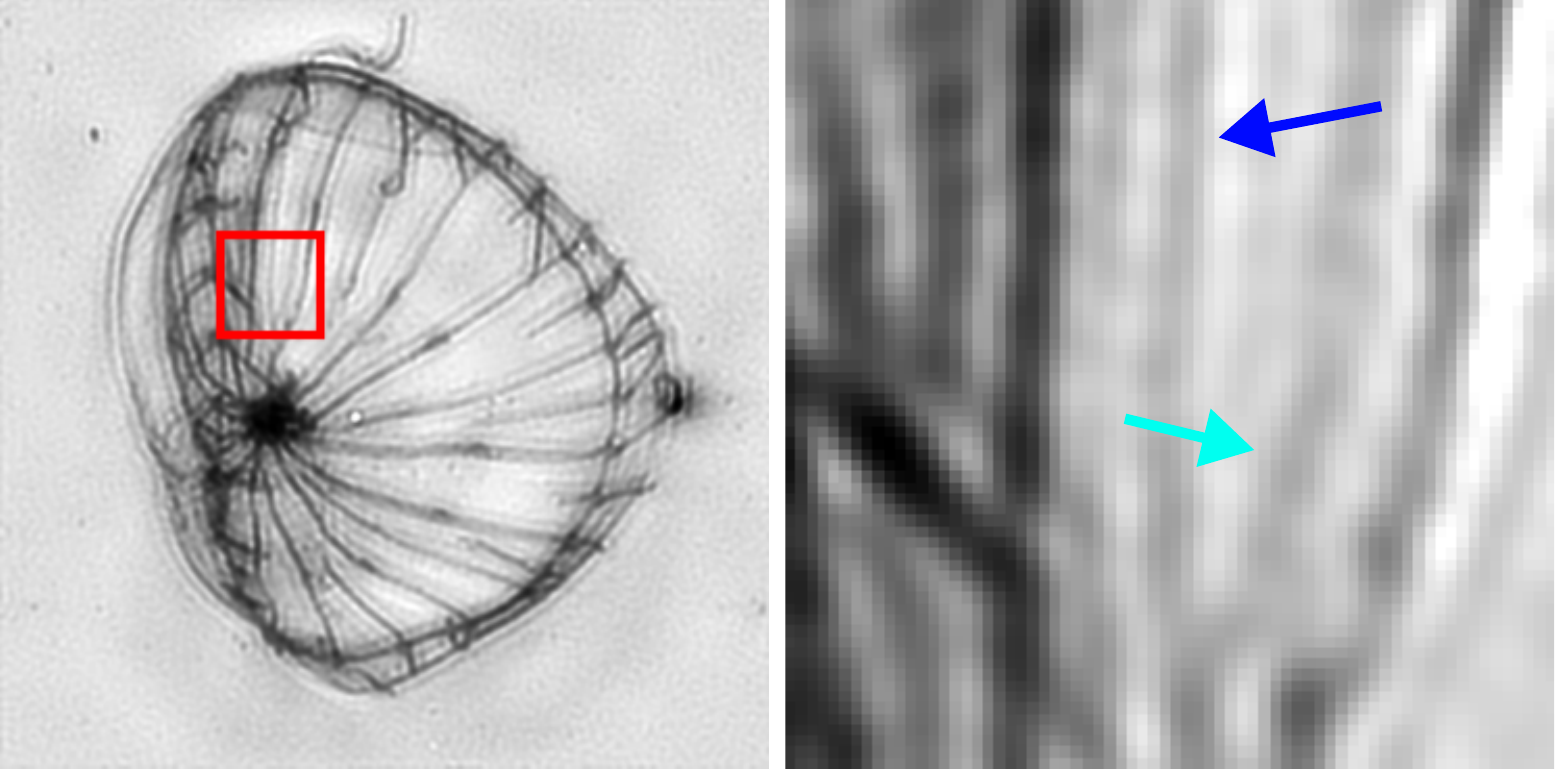}
        \caption{}
    \end{subfigure}
    \\
    \begin{subfigure}[t]{0.45\textwidth}
        \centering
        \includegraphics[width=\linewidth]{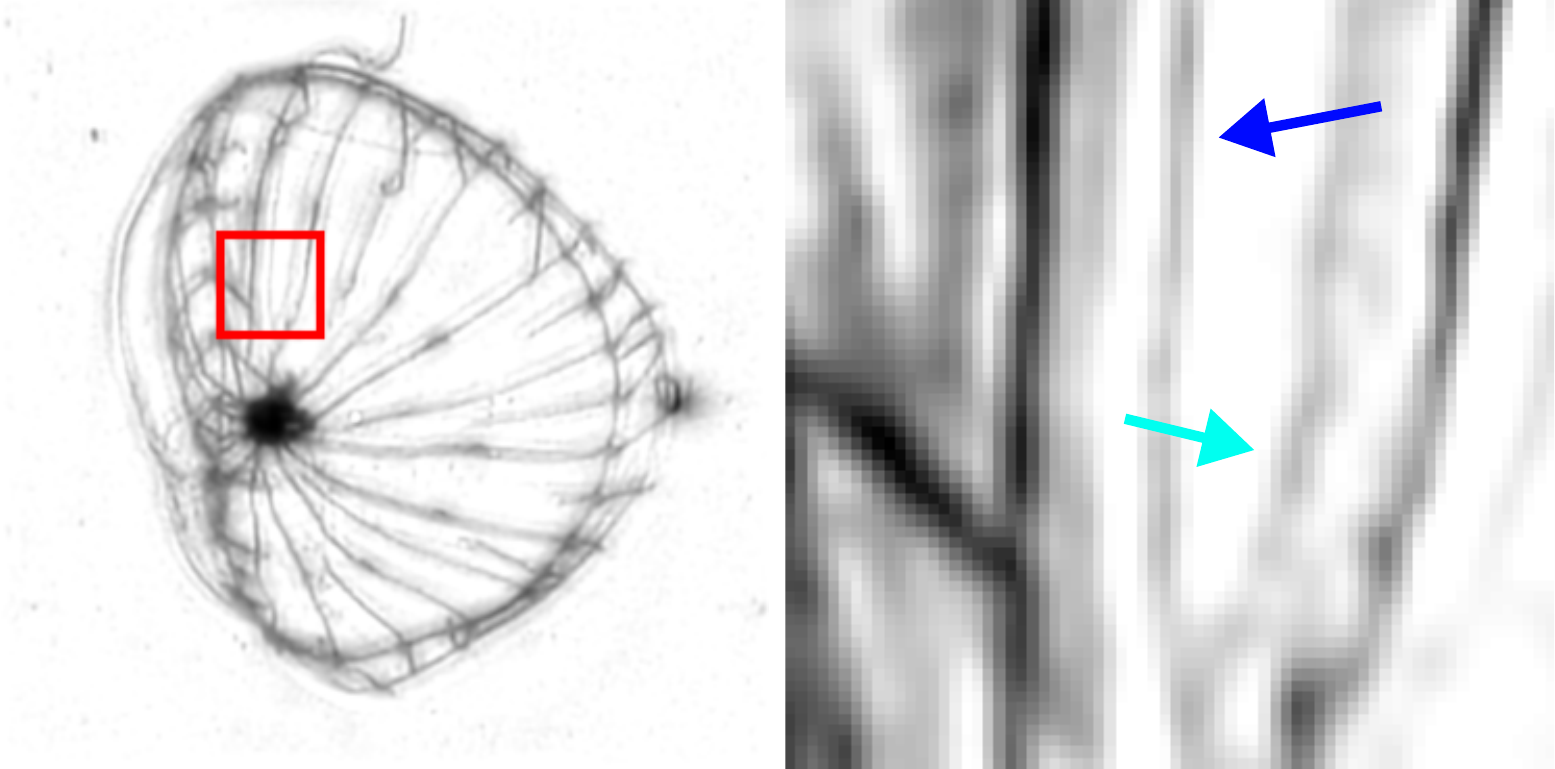}
        \caption{}
    \end{subfigure}
    \caption{Visual comparison involving three images: (a) the raw image captured by an underwater microscope, (b) the deconvolution of a background-subtracted image using the classic Richardson-Lucy algorithm \cite{Richardson:72}, and (c) the image restoration result obtained using CiDeR. The result from CiDeR exhibits improved visual features and fidelity.}
    \label{fig:edof}
\end{figure}

In Figure \ref{fig:edof}, we present a comparison between the classic Richardson-Lucy algorithm \cite{Richardson:72} and CiDeR. The raw image obtained by an underwater microscope exhibits low contrast, and many features of the biological object remain obscured. After restoration, both methods enhance the visibility of the features, but CiDeR outperforms the Richardson-Lucy algorithm in significantly improving the clarity of the object. With the ability of CiDeR to maintain object structures while elevating the overall image quality, it demonstrates its potential as a robust and effective solution for microscopy image restoration tasks.

\subsection{Results with Simulated Blur}

\begin{table}[]
    \centering
    \vspace{2mm}
    \begin{tabular}{ccc}
        \toprule
        Image   & SelfDeblur \cite{selfdeblur}  & CiDeR (ours) \\
        \midrule
        1       & 0.8619                        & \textbf{0.9108} \\
        2       & 0.8935                        & \textbf{0.9043} \\
        3       & 0.9026                        & \textbf{0.9174} \\
        4       & 0.7876                        & \textbf{0.8496} \\
        \midrule
        Average    & 0.8614                        & \textbf{0.8956} \\
        \bottomrule
    \end{tabular}
    \caption{Quantitative comparison of CiDeR and SelfDeblur \cite{selfdeblur} on the dataset introduced by Levin \textit{et al.} \cite{levin}, employing the structural similarity index (SSIM) as the metric. For each row, the value represents the average per image over 8 different degradation models.}
    \label{tab:ssim-levin}
\end{table}

\begin{table}[]
    \centering
    \vspace{2mm}
    \begin{tabular}{ccc}
        \toprule
        Image   & SelfDeblur \cite{selfdeblur}  & CiDeR (ours) \\
        \midrule
        1       & 29.10                        & \textbf{30.99} \\
        2       & 30.35                        & \textbf{30.71} \\
        3       & 32.05                        & \textbf{32.92} \\
        4       & 26.73                        & \textbf{27.89} \\
        \midrule
        Average    & 29.56                        & \textbf{30.63} \\
        \bottomrule
    \end{tabular}
    \caption{Quantitative comparison of CiDeR and SelfDeblur \cite{selfdeblur} on the dataset introduced by Levin \textit{et al.} \cite{levin}, employing the peak signal-to-noise ratio (PSNR) as the metric. For each row, the value represents the average per image over 8 different degradation models.}
    \label{tab:psnr-levin}
\end{table}

\begin{table}[]
    \centering
    \vspace{2mm}
    \begin{tabular}{ccc}
        \toprule
        & SelfDeblur \cite{selfdeblur}  & CiDeR (ours) \\
        \midrule
        No. parameters & 2,357,345 & 322,913 \\
        \bottomrule
    \end{tabular}
    \caption{Overview of the number of learnable parameters.}
    \label{tab:parameters}
\end{table}

\begin{figure}[h!]
    \captionsetup[subfigure]{font=scriptsize,labelfont=scriptsize,{justification=centering}}
    \centering
    \begin{subfigure}[t]{0.20\textwidth}
        \centering
        \includegraphics[width=\linewidth]{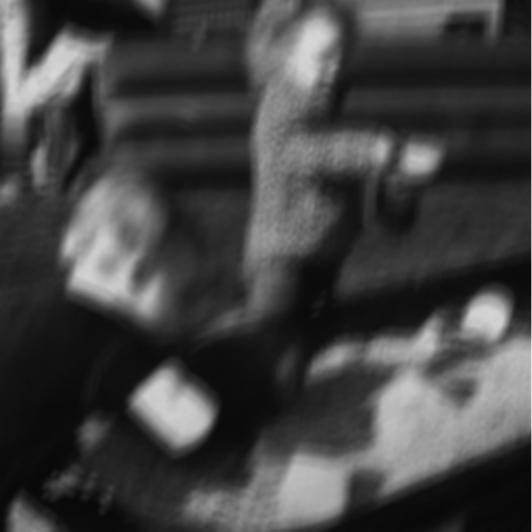}
        \caption{}
    \end{subfigure}
    \begin{subfigure}[t]{0.20\textwidth}
        \centering
        \includegraphics[width=\linewidth]{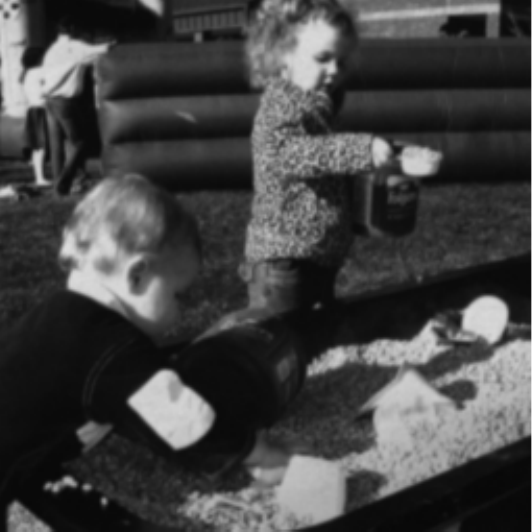}
        \caption{}
    \end{subfigure}
    \\
    \begin{subfigure}[t]{0.20\textwidth}
        \centering
        \includegraphics[width=\linewidth]{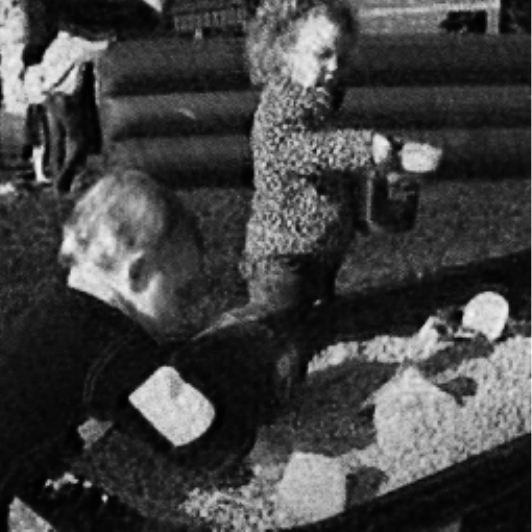}
        \caption{}
    \end{subfigure}
    \begin{subfigure}[t]{0.20\textwidth}
        \centering
        \includegraphics[width=\linewidth]{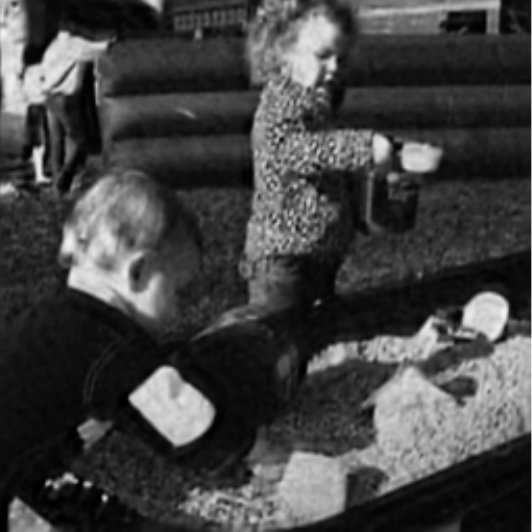}
        \caption{}
    \end{subfigure}
    \caption{Visual comparison of the deconvolution results of an image from the dataset of Levin \textit{et al.} \cite{levin}. Subfigure (a) shows the degraded image, while subfigure (b) displays the ground-truth sharp image. Subfigures (c) and (d) demonstrate the reconstruction results obtained from SelfDeblur \cite{selfdeblur} and CiDeR, respectively.}
    \label{fig:levin}
\end{figure}

We compare CiDeR with SelfDeblur \cite{selfdeblur}, which is the state-of-the-art self-supervised non-blind deconvolution method, using the dataset of Levin \textit{et al.} \cite{levin}. The dataset comprises 32 images, that were generated using 4 sharp images and 8 distinct kernels to simulate various image degradations. Since both models are evaluated in a non-blind setting, we utilise the kernels estimated by \cite{levin} to facilitate the restoration process. Tables \ref{tab:ssim-levin} and \ref{tab:psnr-levin} list the average SSIM and PSNR metrics per each image. In terms of both metrics, CiDeR outperforms SelfDeblur while having significantly less parameters as shown in Table \ref{tab:parameters}.

\section{Discussion \& Future Work}

Image degradation is a common problem in a wide range of field, including telescope imaging, microscopy imaging, or medical imaging. Leveraging the Richardson-Lucy algorithm and DIP assumptions, our image restoration technique proves to be a valuable tool for reconstructing images in these diverse fields, with a potential to significantly improve the accuracy of downstream tasks. In addition to its restoration capabilities, our method benefits from the incorporation of a classic deconvolution algorithm, which results in a substantial reduction in the model size. This reduction proves particularly advantageous in fields where computational resources are scarce, as it significantly lowers the overall computational cost involved in image restoration. As a result, our approach not only produces high-quality reconstructions but also offers an efficient and resource-friendly solution for restoring images, making it better-suited for application in resource-constrained environments compared to existing methods.

Unlike the feature-based Wiener deconvolution model \cite{dong2020deep}, CiDeR takes a different approach by not relying solely on an end-to-end trained feature refinement model. As a result, it effectively addresses the issue of domain gap, since the final image generator is directly optimised on the input image. This direct optimisation mitigates potential challenges associated with domain mismatches, ensuring a more faithful image restoration process. Furthermore, the use of the Wiener filter introduces computational bottlenecks, primarily due to its reliance on the Fourier transform. Such computational costs are particularly evident when dealing with volumetric or 3D data processing \cite{pekurovsky2012p3dfft}. In contrast, using Richardson-Lucy algorithm offers a significant advantage in terms of computational efficiency, especially for volumetric data \cite{li2022incorporating, chobola2023lucyd}. This efficiency opens up exciting possibilities for future work to explore feature-based 2.5D or 3D image restoration tasks, allowing for more extensive and sophisticated applications in three-dimensional data processing.

\section{Conclusion}

In this paper, we propose a novel zero-shot image restoration method called CiDeR that effectively combines a pre-trained feature extraction network with a classic deconvolution algorithm, namely Richardson-Lucy. Our approach leverages deconvolved features to synthesize a reconstructed image, offloading the actual deconvolution task to the classical algorithm. As a result, the self-supervised generator solely focuses on learning the synthesis process, allowing us to design a smaller network compared to existing state-of-the-art methods. This reduction in model size proves beneficial as it conserves computational resources, making our approach particularly well-suited for scenarios where such resources are limited.

Furthermore, our method is designed to be self-supervised, a crucial advantage in biomedical fields where data scarcity is prevalent, and building end-to-end networks is often not possible. Through extensive testing on a classic computer vision benchmark dataset and real microscopy images, the results demonstrate that our method excels in image restoration tasks. This enhanced image quality is particularly crucial for downstream tasks, where accurate images with high fidelity play an important role. Overall, CiDeR exhibits promising potential for various practical applications, which makes it a valuable tool that addresses the challenges of image restoration in domains where data is scarce and computational resources are limited.

\section*{Acknowledgement} 

Tomáš Chobola is supported by the Helmholtz Association under the joint research school "Munich School for Data Science - MUDS".

{\small
\bibliographystyle{ieee_fullname}
\bibliography{egbib}
}

\end{document}